# Text2shape Deep Retrieval Model: Generating Initial Cases for Mechanical Part Redesign under the Context of Case-Based Reasoning

Tianshuo Zang [†], Maolin Yang [†], Wentao Yong and Pingyu Jiang *

State Key Laboratory for Manufacturing Systems Engineering, Xi'an Jiaotong University, Xi'an 710049, China; ibubble@stu.xjtu.edu.cn (T.Z.); maolin@xjtu.edu.cn (M.Y.); yongwentao@stu.xjtu.edu.cn (W.Y.)
* Correspondence: pjiang@mail.xjtu.edu.cn
† These authors contributed equally to this work.

**Abstract:** Retrieving the similar solutions from the historical case base for new design requirements is the first step in mechanical part redesign under the context of case-based reasoning. However, the manual retrieving method has the problem of low efficiency when the case base is large. Additionally, it is difficult for simple reasoning algorithms (e.g., rule-based reasoning, decision tree) to cover all the features in complicated design solutions. In this regard, a text2shape deep retrieval model is established in order to support text description-based mechanical part shapes retrieval, where the texts are for describing the structural features of the target mechanical parts. More specifically, feature engineering is applied to identify the key structural features of the target mechanical parts. Based on the identified key structural features, a training set of 1000 samples was constructed, where each sample consisted of a paragraph of text description of a group of structural features and the corresponding 3D shape of the structural features. RNN and 3D CNN algorithms were customized to build the text2shape deep retrieval model. Orthogonal experiments were used for modeling turning. Eventually, the highest accuracy of the model was 0.98; therefore, the model can be effective for retrieving initial cases for mechanical part redesign.

## 1. Introduction

Case-based reasoning mechanisms can be effectively applied in product redesign activities by retrieving the historical design solutions most similar to new design requirements [1]. The retrieved solutions can then be further modified to fit to the new design requirements [2]. Generally, two types of methods can be applied for the retrieving task, which are manual retrieving and simple reasoning algorithms based-retrieving. However, both of these have disadvantages. For example, the manual retrieving method has the problem of low efficiency when the design solution base is large, and simple reasoning algorithms cannot cover all the structural features in complicated design solutions. For example, when using a decision tree model for a belt pulley design task [3,4], a group of production rules would first be established and then combined into a decision tree. During the process, each of the production rules would consider a functional or structural feature of the belt pulley. Therefore, if there were large numbers of design features to be considered, the decision tree would grow large and complicated, and consequently, the reasoning efficiency would decline.

With the fast development of deep learning techniques, a deep learning-based shape retrieval model has become a possible solution for the problems mentioned above [5–8]. According to the format of the input data, retrieval models can be separated into a multi-dimension retrieval and a single dimension retrieval, and single-dimension retrieval can be further separated into text-based retrieval [9], image-based retrieval [10–15], sketch2shape [16–18], and descriptor2shape [19–22]. Generally, original product design requirements are described in the form of text descriptions, and therefore, text2shape retrieval could be more convenient for design solution retrieval tasks in case-based reasoning-driven product redesign.

However, there are at least two problems to be considered when building a deep learning-based text2shape retrieval model for mechanical part redesign tasks. Firstly, original design requirements are usually described in the form of text descriptions, which are one-dimension data, while the design solutions for mechanical parts are in the form of three-dimensional shapes, and therefore, a data dimension gap must be covered for accurate retrieving from low-dimension data to high-dimension data [23]. On the other hand, the text2shape model for mechanical part retrieving is more complicated than the models for simple shape retrieving tasks (e.g., the commonly studied table, desk, and chair retrieving models) because the structural features of mechanical parts are more complicated than simple shapes, and consequently, the data feature space would be more complicated [24].

In this regard, an integrated deep learning model is established for 3D shape retrieval. The model can retrieve the 3D shapes of mechanical parts according to the input texts that describe new design requirements. In this way, the retrieved 3D shapes can be used as the initial design solutions for the redesign task under the context of case-based reasoning. Our retrieval model has a strong engineering design background, for which we have constructed a specific training set of 1000 samples of linking rods, where each of the samples is a pair consisting of a text description and its corresponding 3D shape model of a mechanical part. Based on the training set, a CNN-RNN and 3D CNN integrated text2shape retrieval model has been established, in which the model structure and parameters are specifically customized to fit the latent space of the training set. It is also worth mentioning that the text descriptions of the samples are organized based on a feature engineering procedure, in which the key structural features of the target mechanical parts (i.e., the linking rods) are analyzed, and therefore, the model can have higher retrieval accuracy compared to free text description-based retrieval. In addition, an orthogonal experiment-based turning method has been applied in order to quickly identify the suitable hyperparameters of the deep learning model.

The rest of the paper is organized as follows. Section 2 reviews the work related to the establishment of our text2shape deep retrieval model. Section 3 describes in detail how we built our training set based on feature engineering and how we built the deep retrieval model to fit to the training set. Section 4 uses a case study to demonstrate the operability of the model in the retrieval tasks of linking rods. Section 5 discusses the contributions of the work and presents a brief conclusion.

**2. Related Work**

As mentioned in Section 1, the core idea of this work is utilizing a text2shape retrieval model to identify the 3D shapes of the historical design solutions most similar to the new design requirements. The design requirements are described in the form of texts. Therefore, text data presentation [25], 3D shape data representation, and text2shape retrieval are the three underlying techniques for the work. In this subsection, we review these three techniques and discuss the correspondent research gap.

*2.1. Text Data Representation and Analysis*

Jiang et al. [26] proposed a deep neural network-based text representation model called topic-based CBOW, and the most important characteristic of this model is its capability for global and local context fusion. Based on the word distribution representation obtained through topic-based CBOW, a short text representation method based on a TF-IWF-weighted pooling mechanism was proposed. Yan et al. [27] proposed a primitive representation learning method that aims to expand internal representations of scene text images. A key point of this method is to model the elements in the feature map as undirected graph nodes. The framework of this method is mainly composed of a pooling aggregator, a weighted aggregator, and a primitive representation learning network. The role of the aggregator is to convert the original representation into a high-level visual text representation. The method shows high efficiency in scene text recognition tasks. Luo et al. [28] proposed an integrated method of Latent Dirichlet Allocation (LDA) text representation and CNN for sentiment analysis in text data. In this method, LDA is used to train the topic distribution of short texts in the latent space, and GRU-CNN is used as the classifier for feature vectors, which improves the accuracy of text classification.

Sinoara et al. [29] proposed an approach to represent document collections according to the embedded representation of words, during which the distance between the pivot word and the context word are calculated to acquire the weights of the words in the context. In this way, an effective way to improve the interpretability of embedded vectors is provided. Song et al. [30] proposed a joint variable autoencoder (VAE) to represent case text embedded representation. This method uses VAE to embed the statistical features and content features of the case texts into the same latent space, and it achieved better performance compared with the models with a single feature embedding mechanism. Shun et al. [31] proposed a polyseme aware vector representation model (PAVRM). The model can generate accurate word vector representations, and it can also identify polysemous words in the corpus through a context clustering algorithm. Hou et al. [32] proposed an Entity-based Concept Knowledge-Aware (ECKA) model, which can integrate semantic information into short text representations. The model was developed based on the CNN algorithm, and it can extract semantic features from words, entities, concepts, and knowledge layers. The model can solve the problem of the poor classification effect caused by the sparsity and shortness of texts.

It can be seen from the literature mentioned above that both DNN and CNN can be effectively used to represent the features of text data. However, in our work, the texts that describe the structural features of mechanical parts are relatively longer, and, consequently, the feature space would be more complicated. Therefore, it could be more effective to use an integrated model rather than a single model to represent the texts that describe the structural features of mechanical parts.

*2.2. 3D shape Data Representation and Analysis*

Feng et al. [33] proposed a mesh natural network, named MeshNet, for 3D shape representation. Shape units of human faces and feature segmentation mechanisms were introduced into the model, thus creating a general model for 3D shape representation for human faces. In this model, a spatial block and a structure descriptor block were established for initial feature representation, and a convolution block was established for aggregating the adjacent features. Gao et al. [34] proposed a kind of local structure-aware anisotropic convolution operation (LSA Conv) for 3D mesh representation learning. The LSA conv operation can learn the adaptive weight matrix of each node, and it does not require pre-processing and post-processing steps. Rinon et al. [35] proposed an unsupervised learning MRGAN algorithm based 3D shape representation method for parts separation. Its network architecture adopts a tree structure diagram, and each branch of its root represents a part with a different shape. The method, by applying a root mixing training strategy and a set of loss functions, can promote branch separation and avoid branch degradation or overgrowth. Jiang et al. [36] proposed a local implicit grid (LIG) method for 3D shape representation, and an implicit eigenvector coding mechanism was used in each region of the regular grid to solve the problem of scalability and universality. Zheng et al. [37] proposed an unsupervised deep implicit template method for 3D shape representation. The key idea is to express deep implicit functions (DIFS) as conditional deformations of template implicit functions, and space warping LSTM was proposed to decompose the conditional space transformation into multiple point-wise transformations in order to ensure generalization ability. Zhang et al. [38] developed a kind of multi-resolution deep implicit function (MDIF) to characterize the shape characteristics of 3D models. A prominent characteristic of the model is a special hierarchical network structure which is able to represent different levels of 3D shape details, and it can improve the performance of shape generation by involving multi-resolution mechanisms and decoder optimization. Tristan et al. [39] proposed a new shape representation method based on implicit architecture, which can realize fast differential rendering, and thus solves the problem of inefficient rendering in implicit shape representation. Wei et al. [40] proposed a canonical view to represent the shape features of 3D models. The canonical view can convert the original features of any view into a fixed number of view features. In addition, a feature separation constraint of the normalized view was proposed to ensure that the view features in the normalized view representation could be embedded into the scattered points in Euclidean space, and could ensure the discernibility of features. Edgar et al. [41] presented a new mid-level patch-based surface representation. This method was a new application of partial point cloud completion of depth maps, shape interpolation, and object generation models. The patches in the model can represent any topological shape and at the same time support arbitrary resolution extraction. Hao et al. [42] proposed a 3D feature descriptor called a geometric feature statistics histogram (GFSH). The descriptor operates by firstly constructing a weighted covariance matrix to solve the stable and reliable local reference frame (LRF), and then performing statistics on multiple geometric distribution features in order to realize the description of 3D features. This method can mitigate the problem of imbalance among effectiveness, robustness, compactness, and efficiency of 3D feature descriptors.

It can be seen from the literature mentioned above that the calculation process of using complicated deep learning network to directly represent the data structures of 3D shapes or the 2D images of 3D shapes can be complicated and time consuming. Therefore, how to improve the efficiency and accuracy of 3D shape representation at the same time is still a research gap to be covered for its engineering application.

*2.3. Retrieval from Text Descriptions to 3D Shapes*

Han et al. [43] proposed a view-based model called Y2Seq2Seq. The model was based on the joint reconstruction of view sequence and word sequence to learn the cross-modal matching relationships. Through two coupled "Y" type sequence to sequence (Seq2Seq) structures, which are the main structures of the model, the Y2Seq2Seq structures can effectively connect the cross-modal semantics. Yue et al. [44] introduced a tri-modal training scheme for text to 3D shape retrieval, including voxels, images, and texts. By applying multi-view 3D model rendering and an end-to-end training method, the model shows better performance than bi-modal models. Wu et al. [45] proposed a two-stage learning method for multi-modal mapping. The first stage combines low-level features and semantic information to represent cross-modal semantic feature vectors, and the second stage uses an encoder–decoder paradigm to learn the matching relationships. This method can preserve feature information and semantic information by projecting multi-modal data into low-dimensional embedding. Chuan et al. [46] proposed a joint embedding method of 3D point cloud and text data to support the matching between text and 3D shapes, and triplet ranking loss was applied in order to calculate the distance between the 3D shape and the text data. Qiao et al. [14] established a two-level retrieval model for assembly structures. In the first-level retrieval model, semantic retrieval was applied in order to identify the mostly matched structures from a model library. In the second-level retrieval model, a kind of attribute adjacency graph-based geometry structure retrieval mechanism was applied in order to output the final results. Li et al. [47] established a semantic tree-based 3D shape retrieval model. The semantic tree was built with WordNet and Princeton Shape Benchmark ontology library, and supported with it, the model showed better performance in parallel experiments.

It can be seen from the literature mentioned above that multi-modal retrieval does have better permanence than single-modal retrieval. However, it would also reduce the convenience of retrieving the required historical design solutions by asking for more types of searching conditions. On the other hand, most of the research on text2shape retrieval was conducted in the field of computer science, and only a few have been devoted to mechanical part shape retrieving tasks. Therefore, a specific data set and its corresponding deep learning model should be constructed for our specific application scenario.

### 3. The Text2shape Deep Retrieval Model

As mentioned in Section 1, the text2shape retrieval model would be able to identify the historical design solutions most similar to the new design requirements, and the identified design solutions can be used as the initial solutions for product redesign in the context of case-based reasoning. In this section, firstly, feature engineering is applied in order to identify the key structural features of the target mechanical parts (i.e., linking rods), a training set with 1000 samples is built for the text2shape retrieval model, and then an integrated model of RNN and 3D CNN is established to fit to the latent space of the training set. The application scenario and construction steps of the model are shown in Figure 1. Supported with the retrieval model, after entering the text descriptions of the structural features of the new design requirements, a list of 3D shapes most similar to the text descriptions can be retrieved from the historical design solution base, and thus provides the initial design solutions to the designers for further redesign procedures.

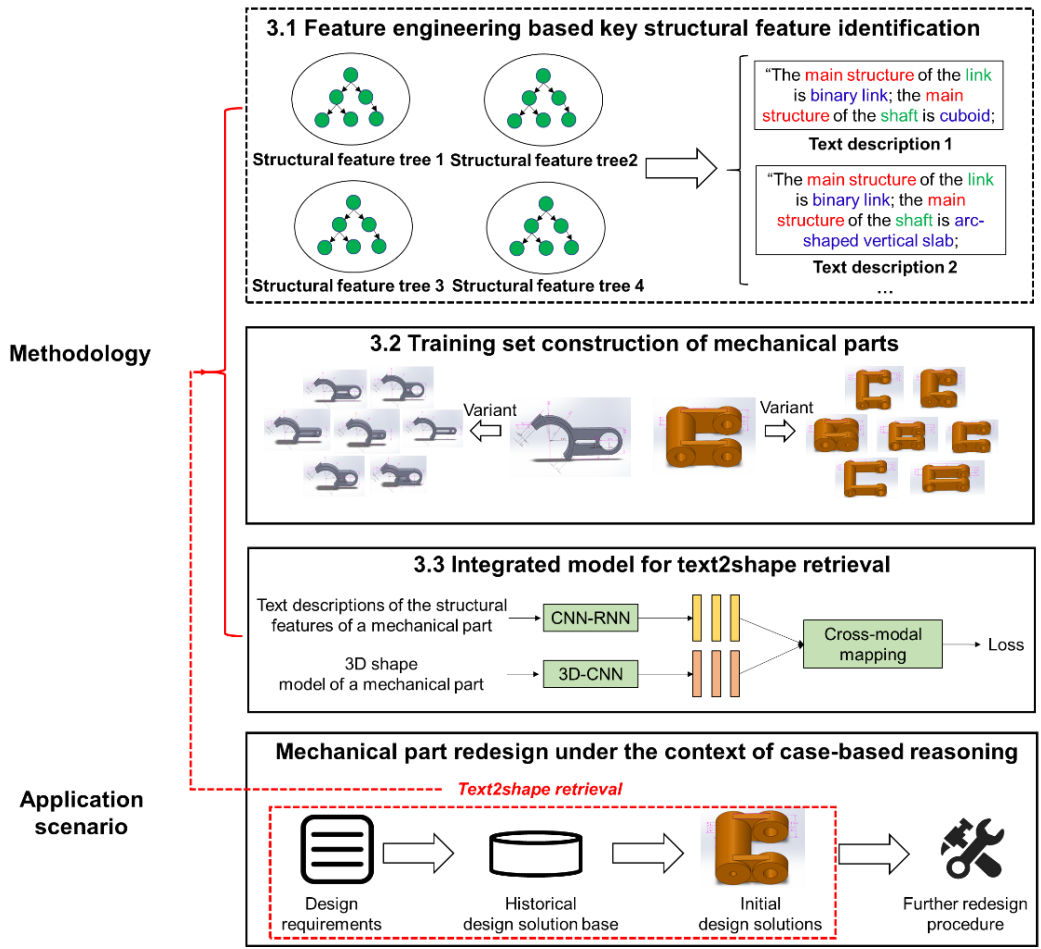

**Figure 1.** The application scenario and construction steps of the text2shape retrieval model.

*3.1. Feature Engineering Based Identification for Key Structural Features*

Feature engineering can be roughly considered as a serious of methods to extract more explicit features from a raw data set and, thus, improve the performances of deep learning models [48,49]. In this subsection, a tree-structure distribution is established in order to reveal the key structural features of linking rods, and a training set would be built accordingly (in the next subsection) to train the text2shape retrieval model. The main steps of feature engineering-based key structural feature identification are shown in Figure 2. It is worth mention that linking rods are used as the target mechanical parts because they are widely used in different types of mechanical assemblies, and there are rich varieties of structural features within them.

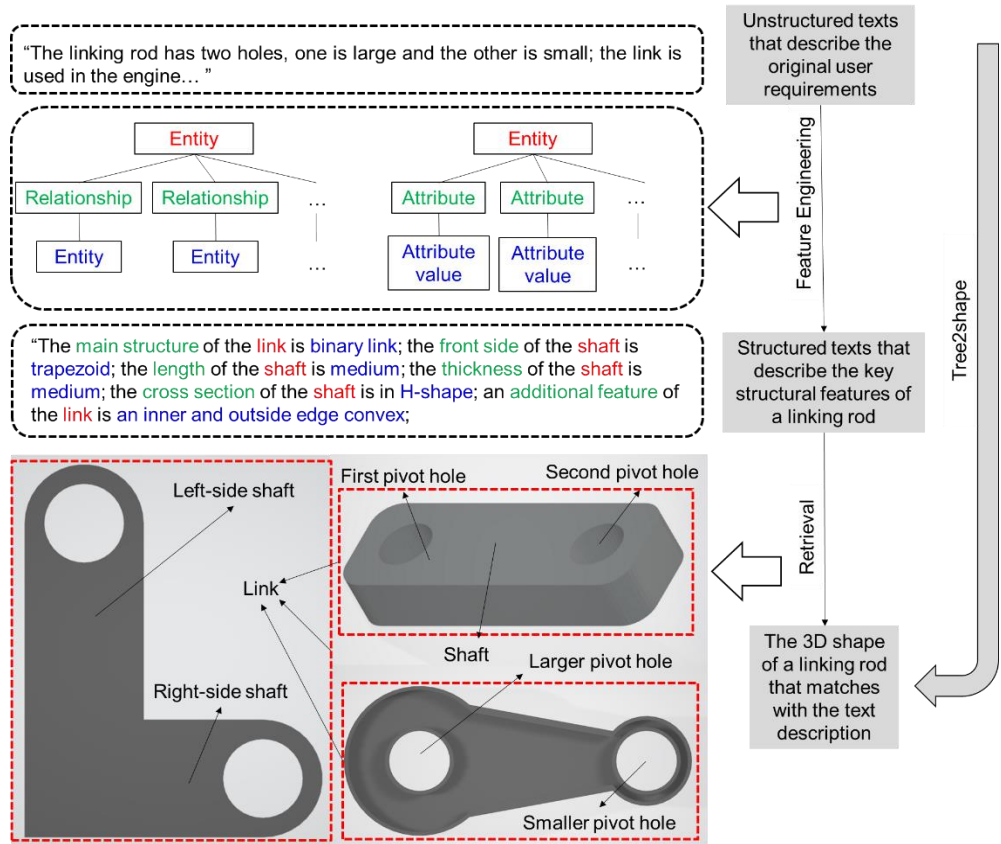

**Figure 2.** The main steps of feature engineering based key structural feature identification.

**Step 1. Key structural feature identification and terminology definition.** The original design requirements expressed by customers are usually unstructured and contain different terms for the same structure features. Therefore, it would largely increase the difficulty for a deep learning model to represent the text data that describe the original design requirements. In this regard, the key structural features of the target mechanical parts should first be identified according to the domain knowledge of professional designers, and the combination of the structural features should be able to describe any variant in the family of the target mechanical parts. In addition, each of the structural features should be provided with exclusive terminology to avoid ambiguity. In this regard, a set of key structural features that define a linking rod has been concluded (e.g., *Binary link*, *Larger pivot hole*, *Inner diameter*), and each of them has been provided with a unified name, as shown in Figure 3.

**Step 2. Triplet-based feature relationship expression.** In this step, the relationships between each pair of structure features are described as "*Entity–Relationship–Entity*," and each structural feature and its possible feature values are described in the form of "*Entity–Attribute–Attribute value*." For example, First pivot hole is a structure-related feature of Connecting rod,; therefore, a piece of triplet can be generated as "*Connecting rod–Structural feature–First pivot hole*." Type is an attribute of First pivot hole, and Separate type could be a possible feature vale of Type; therefore, this piece of structural feature can be expressed as "*First pivot hole–Type–Separate type*." It should be noted that the attributes of a feature can be roughly separated into two types, which are structure-related attributes and size-related attributes. For example, Separated type is a structure-related attribute of First pivot hole, and Inner diameter is a size-related attribute of First pivot hole. It is also worth mentioning that the values of structure-related attributes can also have their own attributes and attribute values. For example, in Figure 3, Additional feature is a structure-related attribute and Inner edge convex is one of its attribute values, while Thickness is a size-related attribute of Inner edge convex, and Thickness also has its attribute values (i.e., *Small*, *Medium*, *Large*).

**Step 3. Tree-structure feature description.** After expressing the key structural features in the form of triplets, they can be connected in the form of tree-structure description, as shown in Figure 3. In total, eight feature entities have been identified (i.e., *First pivot hole*, *Large pivot hole*, *Shaft, Link*, etc.), and each of them can be developed into a tree-structure with their structural-related attributes and size-related attributes. It is worth mentioning that each of these entities can be also connected to the entity of Linking rod with the relationship of "Structural feature."

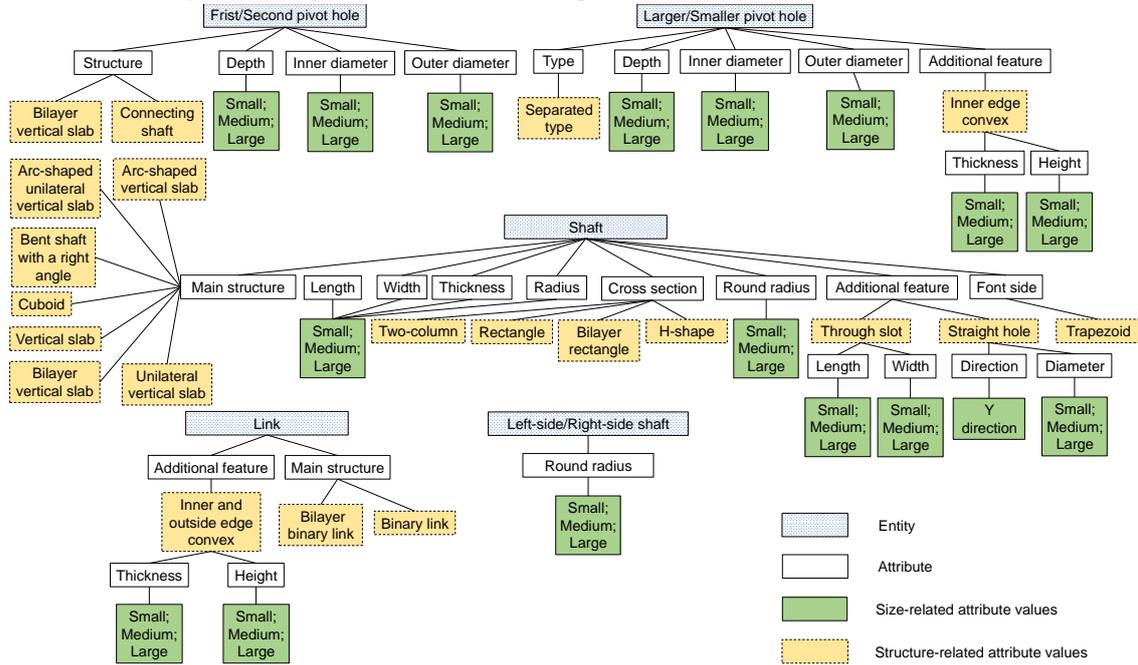

**Figure 3.** The key structural features that determine the 3D shapes of linking rods.

*3.2. Training Set Construction*

In this subsection, a training set with 1000 samples is established for our text2shape deep retrieval model. Each of the samples includes the 3D shape model of a linking rod and a paragraph of texts which describes the key structural features of the linking rod.

3.2.1. Constructing the 3D Shape Models

Firstly, 15 different types of linking rods are selected from different engineering application scenarios, and each of them has different kinds of structure-related feature attributes which make them exclusive from each other. *Solidworks* 2021 was used to draw the original 3D models of the linking rods in the form of STL files. During the process, the sizes of the 3D models were limited within a cubic space of 64 mm3. Then, each of the 3D models were modified into 48 to 64 variant models by changing its size-related feature attributes. For example, by drawing the same type of linking rod with a different shaft length, different variant models of this linking rod can be acquired. However, this work is among the first attempts to retrieve 3D shapes with size-sensitive input conditions; therefore, only three values (i.e., small, medium, large) for the size-related feature attributes were involved in improving the performance of the deep retrieval model.

Secondly, to make it easier for the deep retrieval model to represent the 3D models of the linking rods, the 3D models, in the form of STL files, were then transformed into 3D voxel models, which include both surface voxel information and internal voxel information, and the AABB bounding box method was applied during the process. It is worth mentioning that all of the 3D voxel models should be built in a regular cube space such as 43, 163, or 643 (in this work, 163 is used to find the balance between modeling accuracy and training efficiency). This is to make sure that the 3D voxel models would not distort after being resolved and re-rendered in the deep retrieval model.

Eventually, a total of 1000 variant models of the original 15 linking rods were built in the form of 3D voxel models. A few examples of the models are shown in Figure 4.

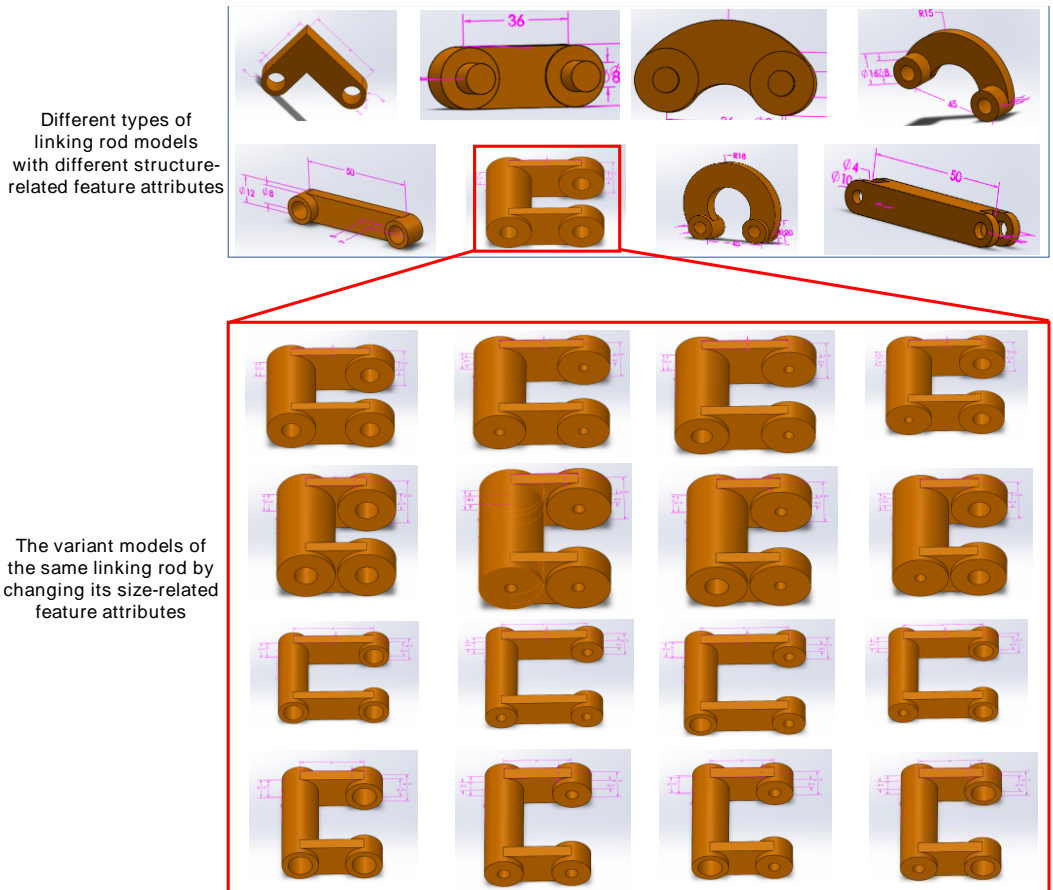

**Figure 4.** Examples of the 3D shape models in the training set.

3.2.2. Preparing the Text Descriptions for the 3D Models

After acquiring the 3D voxel models of the linking rods, each model was provided with a paragraph of text to describe its key structural features. Each paragraph of text description introduced both the structure-related feature attributes (e.g., Type of the linking rod, Main structure of the shaft, Cross-section of the shaft) and the size-related feature attributes (e.g., Inner diameter of the first/second pivot hole, Length of the shaft). It is also worth mentioning that the sentences in the paragraph were expressed in the form of "*Entity–Attribute–Attribute values*." For example, a paragraph of text description could be "The main structure of the link is binary link; the main structure of the shaft is cuboid; the length of the shaft is large; the width of the shaft is medium; the thickness of the shaft is medium". Based on the key structural features identified in Section 3.1 (as shown in Figure 3), any linking rods within the range of the 1000 samples can be accurately defined, and a few examples for the text descriptions in the training set are shown in Figure 5.

| | |
|---|---|
| AAJ002 | The main structure of the link is binary link; the main structure of the shaft is arc-shaped vertical slab; the radius of the shaft is large; the length of the shaft is medium; the thickness of the shaft is large; the cross section of the shaft is rectangle; the inner diameter of the first pivot hole is large; the outer diameter of the first piovt hole is large; the deepth of the first pivot hole is large;  the inner diameter of the second pivot hole is large; the outer diameter of the second piovt hole is large; the deepth of the second pivot hole is large. |
| AAJ003 | The main structure of the link is binary link; the main structure of the shaft is arc-shaped vertical slab; the radius of the shaft is medium; the length of the shaft is large; the thickness of the shaft is large; the cross section of the shaft is rectangle; the inner diameter of the first pivot hole is large; the outer diameter of the first piovt hole is large; the deepth of the first pivot hole is large;  the inner diameter of the second pivot hole is large; the outer diameter of the second piovt hole is large; the deepth of the second pivot hole is large. |
| AAJ004 | The main structure of the link is binary link; the main structure of the shaft is arc-shaped vertical slab; the radius of the shaft is medium; the length of the shaft is medium; the thickness of the shaft is large; the cross section of the shaft is rectangle; the inner diameter of the first pivot hole is large; the outer diameter of the first piovt hole is large; the deepth of the first pivot hole is large;  the inner diameter of the second pivot hole is large; the outer diameter of the second piovt hole is large; the deepth of the second pivot hole is large. |
| AAJ005 | The main structure of the link is binary link; the main structure of the shaft is arc-shaped vertical slab; the radius of the shaft is large; the length of the shaft is large; the thickness of the shaft is large; the cross section of the shaft is rectangle; the inner diameter of the first pivot hole is large; the outer diameter of the first piovt hole is large; the deepth of the first pivot hole is medium;  the inner diameter of the second pivot hole is large; the outer diameter of the second piovt hole is large; the deepth of the second pivot hole is medium. |
| AAJ006 | The main structure of the link is binary link; the main structure of the shaft is arc-shaped vertical slab; the radius of the shaft is large; the length of the shaft is medium; the thickness of the shaft is large; the cross section of the shaft is rectangle; the inner diameter of the first pivot hole is large; the outer diameter of the first piovt hole is large; the deepth of the first pivot hole is medium;  the inner diameter of the second pivot hole is large; the outer diameter of the second piovt hole is large; the deepth of the second pivot hole is medium. |

**Figure 5.** Examples of the texts that describe the structural features of the linking rods, which can be expressed in the form of "*Entity–Attribute–Attribute value.*"

3.2.3. Setting up a Guidance for Building the Training Set by Multiple Participants

As is known to all, a decent training set is the precondition of a successful deep learning model, but manually building a large training set is tiring and time consuming. Therefore, the 3D shape models and their corresponding text descriptions in this work were prepared by multiple participants, and a few rules were applied during the process in order to avoid mistakes that could be harmful to the performance of the deep retrieval model trained using the training set.

(a) Consistent terminology: Using different terms to express the same features in the text descriptions would increase the number of training samples required for the training. Therefore, all the participators should use the same terminology to express the features.

(b) Using the same layout directions for the 3D models: When providing the text descriptions, the 3D models of the linking rods should be located in the same coordinate system with the same layout directions. For example, in order to accurately describe the direction of a hole structure in the shaft (i.e., X, Y, or Z direction), the pivot holes of all the linking rods should be facing to the same direction.

(c) Unified distinction of pivot holes: For the linking rods of which the pivot holes have the same inner diameter, Left-side/Right-side pivot holes were used as the unified terms to further describe the other structural features of the two pivot holes. For the linking rods of which the pivot holes have different inner diameters (these types of linking rods are usually used on internal combustion engines), Larger/Smaller pivot holes were used as the unified terms.

(d) Unified standards for size-related attribute values: As mentioned before, in order to reduce the difficulty of training the deep retrieval model, the size-related attribute values were simplified as Small/Medium/Large, and the values were given using the average values of the samples as benchmarks.

(e) Unified final check: After the training set has been constructed, the 3D models and their corresponding text descriptions provided by multiple participants should be checked by one designer.

*3.3. An Integrated Model of CNN and RNN for Text2shape Retrieval*

In this subsection, an integrated model of RNN and CNN is established to learn the matching relationships between the 3D voxel models and their corresponding text descriptions. The entire model can be concluded into three parts, namely CNN-RNN-based text data representation (extended from the work in [50,51]), 3D CNN-based 3D voxel data representation, and triplet loss-based similarity calculation. The overall structure of the model is shown in Figure 6.

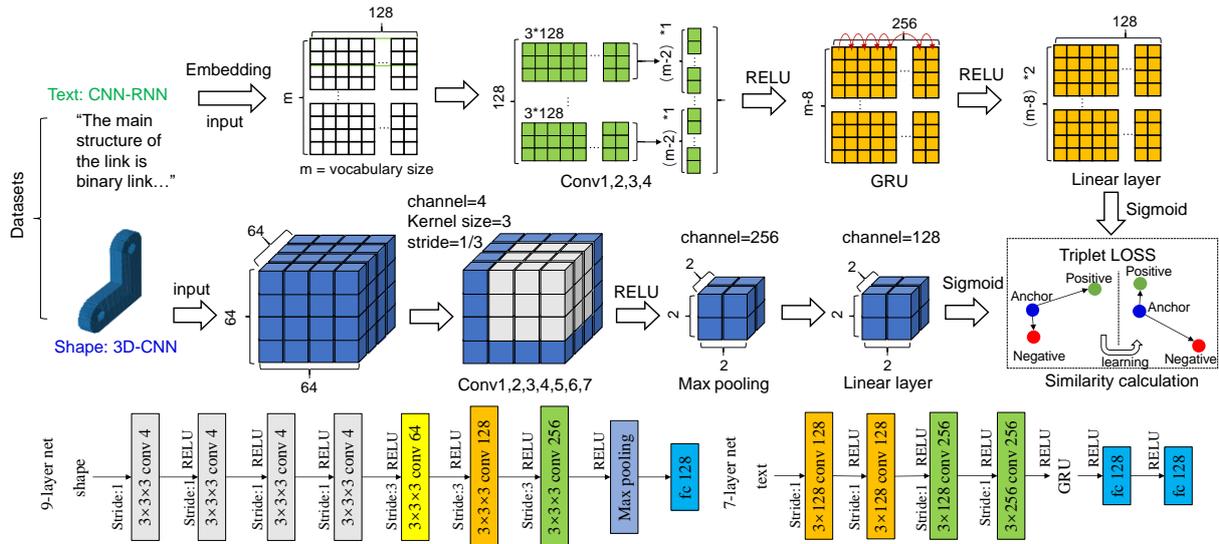

**Figure 6.** The overall structure of the text2shape deep retrieval model.

3.3.1. Representing the Data Features of the Text Descriptions with an Integrated Model of CNN and RNN

An integrated network of CNN and RNN is used to represent the text data that describe the structural features of linking rods into matrices. The structure of the integrated network is shown in the bottom right of Figure 6. It can be seen that the network includes four convolution layers, and each of the convolution layers is followed with a RELU activation function to avoid overfitting. For the first three convolution layers, the sizes of their convolution kernels are $3 \times 128$, the stride is 1, and the padding is 1. For the fourth convolution layer, the convolution kernel size is $3 \times 256$, and the stride and padding are the same as the previous three convolution layers. A Gate recurrent unit (GRU) layer is added between the fourth convolution layer and the fully connected layerin order to obtain the benefits of both RNN and CNNs. The data dimension has been increased to 256 after the fourth convolution layer, and the GRU layer is able to transfer the data into a sequence with temporal dependency among the data. Finally, two fully connected layersare attached at the end of the GRU layer, and the number of channels in the final output is 128.

3.3.2. Representing the Data Features of the 3D Voxel Models with a CNN Based Model

A 3D-CNN network is applied to represent the 3D voxel data into matrices, and its main structure is shown at the bottom left of Figure 6. The 3D-CNN network includes seven convolution layers, and each of them is followed with a RELU activation function to avoid overfitting. For the first four convolution layers, the sizes of their convolution kernels are $3 \times 3 \times 3$, the strides are all set as 1, and the number of channels is set as 4. For the fifth, sixth, and seventh convolution layers, the sizes of their convolution kernels are $3 \times 3 \times 3$, the strides are set as 3, and the numbers of channels are 64, 128, and 256, respectively. A maximum pooling layer with a convolution kernel of $2 \times 2 \times 2$ is attached after the seventh convolution layer for down-sampling, and its stride is set as 1. Finally, a fully connected layeris attached after the pooling layer, and the number of channels in the final output is 128.

3.3.3. Calculating the Similarity between Text Descriptions and 3D Voxel Models with a Triplet Loss Based Method

Text2shape retrieval is a task of cross-modal matching from one-dimensional data to three-dimensional data. It can be seen from the examples in Figure 5 that for each piece of training datum, the texts describe multiple different structural features. During the process, firstly, the retrieval model would represent the text description into a vector, and represent the 3D shape into another vector. Then, the retrieval model would increase the distance between the two vectors during the calculation if the 3D shape did not contain the structural features described by the texts, and vice versa.

In order to improve the accuracy of the text2shape retrieval model, two kinds of distances are calculated and integrated, namely the distance from the vector that represents the texts to the vector that represents the 3D shape model, and the distance from the vector that represents the 3D shape model to the vector that represents the texts. In this regard, the loss function applied in the retrieval model is shown below.

$$Loss = Loss_{t2s} + \mu Loss_{s2t} \quad (1)$$

After acquiring the output vectors of the text descriptions and 3D shape models from Sections 3.3.1 and 3.3.2, respectively, triplet loss [52] can be applied in order to evaluate the distances between the input texts and 3D models. Triplet loss can be calculated with the values of Anchor, Positive, and Negative. The calculation would decrease the distance between Anchor and Positive and increase the distance between Positive and Negative. Triplet loss can be calculated with the equation below.

$$L = max\,(d(Anchor, Positive) - d(Anchor, Negative) + margin, 0) \quad (2)$$

There are three types of calculation methods for Triplet loss function, which are *Easy triplet*, *Hard triplet*, and *Semi-hard triplet*. The Easy triplet does not require further optimization calculation, and it should be applied when L = 0, and

$$d(anchor, positive) + margin < d(anchor, negative) \quad (3)$$

The *Hard triplet* should be applied when the distance between Anchor and Positive is large, and

$$d(anchor, negative) < d(anchor, positive) \quad (4)$$

The *Semi-hard triplets* should be applied when Equation (5) can be satisfied. In this kind of situation, the vectors of Anchor and Negative are relatively close to each other, but there is still a margin between them.

$$d\,(anchor, positive) < d(anchor, negative) < d\,(anchor, positive) + margin \quad (5)$$

Eventually, *Semi-hard triplet* is applied in the retrieval model, and during the training process, the distance between each pair of the vectors that represent the texts and the 3D shape was calculated and compared with the others. On this basis, a smaller loss value indicates that the 3D shape model and the texts are more similar to each other, and vice versa.

**4. Case Study**

In order to verify the effectiveness of the works in Section 3, a Python-based deep retrieval model has been developed, and a series of retrieval experiments has been conducted. During the process, orthogonal experiments were used for turning the hyperparameters and the network structure of the deep retrieval model.

*4.1. Operating Environment*

The experiment platform is built on a deep learning workstation with Intel Xeon silver 4216 CPU and GeForce RTX™ 3090 GPU. The operating environment of the workstation is shown in Table 1. Based on the platform, the model built in Section 3.3 has been trained with the dataset built in Section 3.2.

**Table 1.** Operating environment.

| Project | Project Content |
| --- | --- |
| Operating system | Microsoft Windows 10 |
| Python environment | Python 3.9 |
| Virtual environment | Anaconda3-5.2.0 |
| Programming software | PyCharm 2022.1.3 |

*4.2. Training Set Preparation*

The original samples collected for the purpose of building the training set include a group of text descriptions and their corresponding 3D shape models. The lengths of the original text descriptions were different, as they were manually provided by the designers according to their understanding of the structural features contained in the corresponding 3D shapes; the original 3D shapes generated from *Solidworks* were in the format of *STL* files.

Based on the original samples, a preprocessing procedure was conducted before using the samples to train the retrieval model. During the process, a piece of self-developed code was used to transfer the 3D shapes from STL format to 3D voxel models in *NRRD* format. *NRRD* format was applied as it can store information on dimension, sizes, space directions, space origin, etc., and it can be conveniently processed with plug-in packages in Python. Each of the 3D voxel models was built in a regular cube space of $16^3$. For text descriptions, self-developed codes were used to cut down the length of the texts in each sample to a maximum of 256 words. In addition, the words that appeared more than twice were selected as vocabularies for the training of the text representation model, and all of the text descriptions were stored in *CSV* format.

After the preprocessing, a total of 1000 samples were fabricated and can be used for the training procedure.

*4.3. Training Details*

4.3.1. Tuning the Deep Retrieval Model with an Orthogonal Experiment-Based Method

In order to efficiently identify the optimal configuration of the hyperparameters and network structure settings, including *Batch size*, *Convolution layer number*, *Learning rate*, and *Epoch*, an orthogonal experiment-based tuning method was applied.

Firstly, an orthogonal experiment with four factors and three levels was conducted, as shown in Table 2. It can be found from the results that the accuracy was mainly related to *Batch size*, *Convolution layer number*, and *Learning rate,* because changing the settings of these factors would effectively influence the accuracy. Furthermore, it can be found that a configuration of a large Batch size, a higher Convolution layer number, and a lower Learning rate leads to higher accuracy. The highest accuracy in the first orthogonal experiment was 71.09%.

Table 2. The first orthogonal experiment.

| Number | Batch Size (A) | Learning Rate (B) | Epoch (C) | Convolution Layer Number (D) | Accuracy (E) /% |
|---|---|---|---|---|---|
| 1 | 16 | 0.001 | 50 | 3 | 3.77 |
| 2 | 16 | 0.0001 | 200 | 4 | 64.77 |
| 3 | 16 | 0.00001 | 100 | 5 | 46.45 |
| 4 | 32 | 0.001 | 200 | 5 | 30.00 |
| 5 | 32 | 0.0001 | 100 | 3 | 55.62 |
| 6 | 32 | 0.00001 | 50 | 4 | 69.38 |
| 7 | 64 | 0.001 | 100 | 4 | 48.44 |
| 8 | 64 | 0.0001 | 50 | 5 | 71.09 |
| 9 | 64 | 0.00001 | 200 | 3 | 57.81 |
| $T_{1j}$ | 114.99 | 82.21 | 144.24 | 117.2 | |
| $T_{2j}$ | 155 | 191.48 | 150.51 | 182.59 | 447.33 |
| $T_{3j}$ | 177.34 | 173.64 | 152.58 | 147.54 | |
| $R_j$ | 62.35 | 109.27 | 8.34 | 65.39 | |
| Order | | | B > D > A > C | | |
| Best level | A3 | B2 | C3 | D2 | |
| Best combination | | | A3B2C3D2 | | |

Based on the result of the first orthogonal experiment, the second experiment was designed to expand the range of each factor. The second experiment used four factors and four levels, and the range is larger than the first one, as shown in Table 3. A total of 16 groups of experiments were conducted, and the highest accuracy was improved to 95% when the *Batch size* was 4, the *Convolution layer number* was 7, and the Learning rate was 0.000001.

Table 3. The second orthogonal experiment.

| Number | Learning Rate (A) | Batch Size (B) | Convolution Layer Number (C) | Epoch (D) | Accuracy (E) /% |
|---|---|---|---|---|---|
| 1 | 0.001 | 4 | 4 | 50 | 64.44 |
| 2 | 0.001 | 16 | 5 | 100 | 48.86 |
| 3 | 0.001 | 32 | 6 | 200 | 61.25 |
| 4 | 0.001 | 64 | 7 | 500 | 42.19 |
| 5 | 0.0001 | 4 | 5 | 142 | 53.89 |
| 6 | 0.0001 | 16 | 4 | 500 | 42.61 |
| 7 | 0.0001 | 32 | 7 | 50 | 48.12 |
| 8 | 0.0001 | 64 | 6 | 100 | 58.59 |
| 9 | 0.00001 | 4 | 6 | 142 | 90.00 |
| 10 | 0.00001 | 16 | 7 | 200 | 58.52 |
| 11 | 0.00001 | 32 | 4 | 100 | 82.50 |
| 12 | 0.00001 | 64 | 5 | 50 | 60.16 |
| 13 | 0.000001 | 4 | 7 | 100 | 95.00 |
| 14 | 0.000001 | 16 | 6 | 50 | 85.23 |
| 15 | 0.000001 | 32 | 5 | 500 | 74.37 |
| 16 | 0.000001 | 64 | 4 | 200 | 69.53 |
| $T_{1j}$ | 216.74 | 303.33 | 259.08 | 257.95 | |
| $T_{2j}$ | 203.21 | 235.22 | 237.28 | 284.95 | 1035.26 |
| $T_{3j}$ | 291.18 | 266.24 | 295.07 | 243.19 | |
| $T_{4j}$ | 324.13 | 230.47 | 243.83 | 249.17 | |
| $R_j$ | 120.92 | 72.86 | 57.79 | 41.76 | |
| Order | | | A > B > C > D | | |
| Best level | A4 | B1 | C3 | D2 | |

| Number | Learning Rate (A) | Batch Size (B) | Convolution Layer Number (C) | Epoch (D) | Accuracy (E) /% |
|---|---|---|---|---|---|
| Best combination | | | A4B1C3D2 | | |

Based on the results of the second experiment, the third orthogonal experiment was conducted to further explore the optimal configuration of the hyperparameters and network structure setting. In this experiment, four factors and two levels were applied, as shown in Table 4, and this experiment was conducted to ensure that the ignored configurations of factors in the previous experiments can be covered. It can be seen from the results that the highest accuracy reaches 98% when the *Batch size* is 4, the number of *Convolution layer number* is 7, and the *Learning rate* is 0.00001.

Table 4. The third orthogonal experiment.

| Number | Learning Rate (A) | Batch Size (B) | Convolution Layer Number (C) | Accuracy (D) /% |
|---|---|---|---|---|
| 1 | 0.00001 | 4 | 6 | 91.67 |
| 2 | 0.00001 | 4 | 7 | 77.78 |
| 3 | 0.00001 | 16 | 6 | 98.30 |
| 4 | 0.00001 | 16 | 7 | 59.66 |
| 5 | 0.000001 | 4 | 6 | 96.67 |
| 6 | 0.000001 | 4 | 7 | 97.78 |
| 7 | 0.000001 | 16 | 6 | 94.89 |
| 8 | 0.000001 | 16 | 7 | 86.93 |
| $T_{1j}$ | 327.41 | 363.9 | 381.53 | 703.68 |
| $T_{2j}$ | 376.27 | 339.78 | 322.15 | |
| $R_j$ | 48.86 | 24.12 | 59.38 | |
| Order | | C > A > B | | |
| Best level | A2 | B1 | C1 | |
| Best combination | | A2B1C1 | | |

4.3.2. Verifying the Experiment Results with Range Analysis and Variance Analysis

In this subsection, *Range* and *Variance* values of the orthogonal experiments are analyzed to ensure that the experiment results were not influenced by experiment errors [53]. Here, the results from the third experiment are analyzed for demonstration.

When analyzing the *Range* values, the extreme values can be acquired from *MiniLab* software, and it can be seen that the *Delta* value of the factor of Network layer was the largest (as shown in Table 5), which was consistent with the results in Table 3, and all the three factors were in the same order. Here, *Delta* is a parameter that indicates the response level to the average values, and a larger *Delta* value indicates a stronger influence.

Table 5. Response level to the average values.

| Level | Learning Rate (A) | Batch Size (B) | Convolution Layer Number (C) |
|---|---|---|---|
| 1 | 81.85 | 90.97 | 95.38 |
| 2 | 94.07 | 84.94 | 80.54 |
| Delta | 12.22 | 6.03 | 14.84 |
| Order | 2 | 3 | 1 |

Supported with this Figure 7, the average value of each factor can be easily analyzed. The abscissa of the coordinate indicates the level of each factor, and the ordinate indicates the average values of the average values. It can be seen from the figure that the average value of the average value is the largest when *Learning rate* is at the second level, Batch size is at the first level, and the *Convolution layer number* is at the first level.

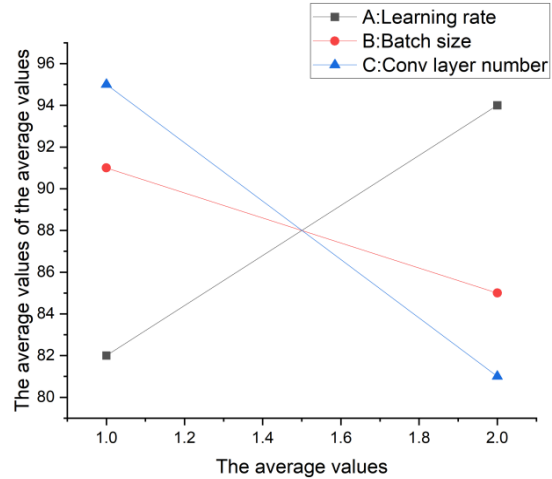

**Figure 7.** The main effect plot of average values.

It can be seen that the configuration of *A2B1C1* gained the highest average value according to the results of the above analysis. However, Variance analysis is still required to ensure that the three factors are statistically significant for the experiment results. Supported by *MiniLab* software, the *Degrees* of freedom, *Adj SS, Adj MS*, and F values can be acquired (as shown in Table 6). According to the F distribution diagram, it can be seen that the above values indicate that factors A and C would have prominent influence when the level is 0.25.

**Table 6.** Variance analysis.

| Factor | Freedom | Adj SS | Adj MS | F value | *p* Value |
|---|---|---|---|---|---|
| A | 1 | 298.41 | 298.41 | 2.75 | 0.173 |
| B | 1 | 72.72 | 72.72 | 0.67 | 0.459 |
| C | 1 | 440.75 | 440.75 | 4.06 | 0.114 |
| Error | 4 | 434.70 | 108.68 | | |
| Total | 7 | 1246.58 | | | |

Eventually, it can be seen that results from the orthogonal experiments are correct according to the analysis on the *Range values* and *Variance values*.

*4.4. Retrieval Case Study*

The case study described in this subsection tested whether the retrieval model trained above could identify the 3D shape models most similar to the input text descriptions. The trained model should be able to identify eight 3D shape models with the smallest loss values to the input text descriptions. More than twelve retrieval experiments were conducted, and all of them output decent results. The input and output of one of the experiments are shown in Figure 8.

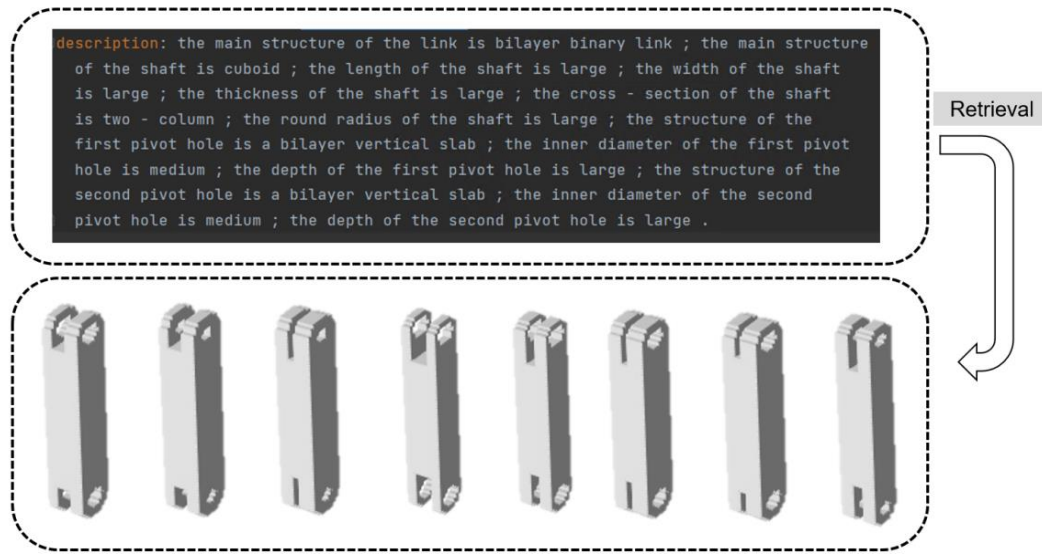

**Figure 8.** An example of the retrieval experiments.

## 5. Discussion and Conclusions

In this paper, a text2shape deep retrieval model has been developed following the steps of feature engineering-based key structural feature identification, training set building, and deep retrieval model building. The contribution of the work can be concluded from two perspectives.

From a methodology perspective, the text2shape deep retrieval model can support acquiring the design solutions most similar to new design requirements from the historical design solution base, and thus, it can provide initial design solutions for redesign tasks under the context of case-based reasoning. Furthermore, the text2shape deep retrieval model-based approach has higher retrieving efficiency and accuracy compared with the commonly used manual retrieval approach and simple reasoning algorithm-based approaches.

From a technology perspective, a specific text2shape matching training set has been established using a linking rod as the retrieval target, and an integrated model of RNN and CNN has been established to fit to the latent space of the training set. In order to avoid the problem of overfitting, the network structure of the retrieval model has been modified. Several convolution layers were added in the 3D-CNN model, and each convolution layer is followed by a RELU function. In contrast to the research in computer science, the training set used in this paper was constructed for this specific application scenario, and the structures and parameters of the deep retrieval model have been modified to fit the latent space of the data feature hidden in the specific training set. In addition, an orthogonal experiment-based method has been applied for model tuning, which helps to quickly identify the most suitable configuration of hyperparameters and network settings for the deep retrieval model and the dataset.

It is worth mentioning that the work in this paper is actually the first step of our attempt to utilize deep learning models to support product design activities. The main reason for expressing the structural features in the form of triplet data (as demonstrated in Section 3.1) is for the next step of our future work, in which the text2shape retrieval model would be replaced by a graph2shape generation model, as a knowledge graph can visually expresses the overall structure of the target product, and the generation model (rather than retrieval model) would be able to generate new design solutions according to the input graph description of the design requirements. The new design solutions generated from the model would also be different from the existing design solutions in the historical case base. In addition, the matching relationship between function requirements and 3D shapes would also be considered in our future work by constructing new training sets and deep learning models that match with the training sets. In the new training sets, the input would be text or graph descriptions of functional requirements, and the output would be the corresponding 3D shapes. In this way, the retrieval/generation from original customer requirements to 3D shapes of the products would be achieved. In addition, it should be noted that more training samples, derived by making variants of different types of linking rods, are still required for improving the generalization capability of the retrieval model.